\documentclass[letterpaper, 10 pt, conference]{ieeeconf}  

\IEEEoverridecommandlockouts                              

\usepackage{graphics} 
\usepackage{epsfig} 
\usepackage{mathptmx} 
\usepackage{times} 
\usepackage{amsmath} 
\usepackage{amssymb}  
\usepackage{siunitx}
\usepackage{pifont}
\usepackage{subcaption}
\usepackage{booktabs} 
\usepackage{multirow}
\usepackage{xcolor}
\usepackage{cite}  
\newcommand{\cmark}{\ding{51}}%
\newcommand{\xmark}{\ding{55}}%

\definecolor{baseline}{rgb}{0.8901960784313725, 0.10196078431372549, 0.10980392156862745}
\definecolor{source}{rgb}{1.0, 0.4980392156862745, 0.0}
\definecolor{examplefreeind}{rgb}{0.45, 0.7, 1.0}
\definecolor{examplefreeseq}{rgb}{0.12156862745098039, 0.47058823529411764, 0.7058823529411765}
\definecolor{joininde}{rgb}{0.55, 0.85, 0.35} 

\definecolor{joinseq}{rgb}{0.2, 0.6274509803921569, 0.17254901960784313}

\DeclareMathAlphabet{\mathcal}{OMS}{cmsy}{m}{n}
\title{\LARGE \bf
	Fine-Tuning Strategies for Continual Online EEG Motor Imagery Decoding: Insights from a Large-Scale Longitudinal Study
}

\author{Martin Wimpff$^{1}$, Bruno Aristimunha$^{2, 3}$, Sylvain Chevallier$^{2}$, and Bin Yang$^{1}$
	\thanks{This research was funded by the Quantum Human Machine Interfaces (QHMI) project within the QSens - Quantum Sensors of the Future Cluster grant number 03ZU1110DC. BA and SC is supported by ANR-22-CE33-0015-01 and ANR-17-CONV-0003.}
	\thanks{$^{1}$Martin Wimpff and Bin Yang are with the Institute of Signal Processing and System Theory,
		University of Stuttgart, 70569 Stuttgart, Germany
		{\tt\small martin.wimpff@iss.uni-stuttgart.de}}%
	\thanks{$^{2}$Bruno Aristimunha and Sylvain Chevallier are with the Inria TAU team, LISN-CNRS, Universite Paris-Saclay
		91400 Orsay, France.
		$^{3}$ Federal University of ABC, Santo Andre, Brazil
	}%
}

\begin{document}

	\maketitle
	\thispagestyle{empty}
	\pagestyle{empty}

	\begin{abstract}
		This study investigates continual fine-tuning strategies for deep learning in online longitudinal electroencephalography (EEG) motor imagery (MI) decoding within a causal setting involving a large user group and multiple sessions per participant. 
		We are the first to explore such strategies across a large user group, as longitudinal adaptation is typically studied in the single-subject setting with a single adaptation strategy, which limits the ability to generalize findings.
		First, we examine the impact of different fine-tuning approaches on decoder performance and stability.
		Building on this, we integrate online test-time adaptation (OTTA) to
		adapt the model during deployment, complementing the effects of prior fine-tuning. 
		Our findings demonstrate that fine-tuning that successively builds on prior subject-specific information improves both performance and stability, while OTTA effectively adapts the model to evolving data distributions across consecutive sessions, enabling calibration-free operation.
		These results offer valuable insights and recommendations for future research in longitudinal online MI decoding and highlight the importance of combining domain adaptation strategies for improving BCI performance in real-world applications.\newline
		{\textbf{\textit{Clinical Relevance}}}\textemdash Our investigation enables more stable and efficient long-term motor imagery decoding, which is critical for neurorehabilitation and assistive technologies.
		
	\end{abstract}

	\section{INTRODUCTION}
	A brain-computer interface (BCI) measures brain activity and translates it into control commands for computers or other external devices \cite{peksa2023state}. 
	This provides a direct alternative to natural neural pathways, enabling BCIs to replace, restore, enhance, supplement, or improve the brain's interaction with its external or internal environment \cite{cervera2018brain, bcisociety}.  
	
	A widely used method for controlling BCIs is the motor imagery (MI) paradigm. 
	In this paradigm, the user imagines the movement of a body part without physically performing the action. 
	This imagined movement engages neural mechanisms similar to those involved in actual execution \cite{decety1996neurophysiological}, making MI-BCIs particularly effective for promoting motor recovery in chronic stroke patients \cite{cervera2018brain}. 
	
	
	However, novice users often struggle to elicit the correct brain patterns, a challenge known as BCI inefficiency \cite{sannelli2019large, zhang2020subject, lee2019eeg}, also referred to as BCI illiteracy. 
	Unlike paradigms such as P300 or steady-state visually evoked potentials, which rely on responses to external stimuli, MI requires users to endogenously modulate their brain rhythms, i.e., actively regulate their neural activity which is known to be more challenging. 
	Potential solutions to this BCI inefficiency fall mostly into two categories: either promoting user learning or improving the decoder \cite{sannelli2019large, zhang2020subject}.  
	
	As with almost any skill acquisition process, effective BCI usage depends on practice guided by feedback \cite{gaume2016psychoengineering}. 
	Research even indicates that implicit learning, where users develop skills through self-regulation guided by feedback, may be more effective than explicitly guiding or instructing the user \cite{kober2013learning, jeunet2016standard, corsi2020functional}. 
	To facilitate such implicit learning, closed-loop systems that provide real-time feedback are essential. 
	
	The other possible solution for 
	successful BCI usage is to increase the quality of the decoder. While this has been the subject of a large research effort for several decades~\cite{lotte2018review}, one important -- although less investigated issue -- is 
	the adaptiveness of the decoder. 
	While users must develop the ability to generate the correct brain patterns, the system must also adjust to the evolving neural activity of the users. 
	Decoder adaptation is essential in this process, ensuring that the BCI remains effective despite evolving brain patterns over time. 
	
	Importantly, user learning and decoder adaptation are not independent processes but are tightly interconnected. 
	This dynamic interaction can be conceptualized as a \emph{two-learner problem}, where both the user and the decoder adapt to each other's changing trajectories over time \cite{millan2015brain, muller2017mathematical, perdikis2020brain}. 
	Through mutual learning, they try to find an optimal communication strategy. 
	In longitudinal settings, this interdependence becomes particularly significant as the user's neural patterns may shift substantially over time due to factors such as learning, neuroplasticity, or changes in the environment. 
	
	Research has demonstrated that decoder adaptation can enhance the user's ability to control the BCI, leading to overall performance improvements \cite{perdikis2020brain, rao2024once, jaeger2023cybathlon, perdikis2018cybathlon, forenzo2024continuous, tortora2022neural}. 
	However, determining the optimal strategy for recalibrating decoders over time remains an open question. 
	Current approaches vary primarily in the frequency of recalibration \cite{jaeger2023cybathlon}, and the data composition used for recalibration \cite{rao2024once, forenzo2024continuous}. 
	
	Within the realm of deep learning, the process of gradually adapting to an incoming stream of data from different domains can be described as domain-incremental continual learning \cite{van2022three, wang2024comprehensive}. 
	However, continual learning in MI decoding presents unique challenges compared to its applications in other fields. 
	In traditional settings, continual learning methods prioritize efficient adaptation to new tasks or domains while retaining knowledge of previous ones, thereby avoiding catastrophic forgetting. 

	In MI decoding, the latter consideration is irrelevant, as the data distribution continuously evolves over time, and revisiting previous distributions is neither necessary nor feasible, considering the non-stationary nature of the data. 
	This differs from applications such as automotive systems, where recurring conditions (e.g., varying weather) require continual learning to handle repeated scenarios. 
	However, it is worth mentioning that a certain level of decoder stability tends to benefit user learning in MI decoding \cite{perdikis2018cybathlon}. 
	
	Another distinctive aspect of continual MI decoding across multiple sessions is the potential absence of calibration data for the upcoming (target) session \cite{rao2024once}. 
	As a result, offline adaptation between sessions must rely solely on the most recent data, i.e., the data from previous session(s). 
	
	To address the distribution shift between consecutive sessions, i.e., the most recent session and the upcoming target session, online test-time adaptation (OTTA) \cite{li2023t, wimpff2024calibration} emerges as a viable approach. 
	OTTA leverages the incoming sample-wise stream of unlabeled target data after deployment to adapt the model dynamically to the evolving unknown target distribution. 
	
	While decoder adaptation and continual learning have shown promise, key questions remain about effectively integrating fine-tuning strategies in longitudinal MI decoding across large user groups. The dynamic interplay between user learning and decoder adaptation, alongside EEG’s non-stationary nature and limited target data, demands broader studies beyond single-subject or short-term settings.

	To address these challenges, we systematically investigate continual learning for online MI decoding in a large-scale longitudinal setting. Our contributions are as follows:
	\begin{itemize}
		\item 
		We are the first to investigate deep learning-based continual learning for MI decoding in a longitudinal setting across a large user group (61 subjects).
		\item
		We examine the impact of different fine-tuning strategies on the performance and stability in a realistic causal pseudo-online \cite{carrara2024pseudo} manner. 
		\item 
		We demonstrate the effectiveness of combining offline fine-tuning together with online test-time adaptation to establish a comprehensive, fully adaptive calibration-free decoding framework. This framework effectively leverages new data as it becomes available to adapt the decoder to users' evolving neural patterns, addressing the domain shifts naturally present in biosignals. It not only ensures sustained performance and stability across sessions and subjects but also enables continuous performance improvements.
	\end{itemize}

	\section{MATERIALS \& METHODS}
	\subsection{Data}
	We employ the dataset published by \textit{Stieger et al.} in 2021 \cite{stieger2021continuous}, which includes data from 61 subjects with 7-11 sessions per user. To our knowledge, this is the only publicly available motor imagery (MI) dataset that captures longitudinal user learning within a large population with online feedback \cite{gwon2023review}. In contrast, commonly used BCI datasets, such as those from the BCI competitions \cite{tangermann2012review}, focus on small user groups (typically around 10 subjects) and neglect user learning, as they include only a few sessions (typically $\leq2$). More recent databases \cite{lee2019eeg, dreyer2023large} have expanded to include significantly larger subject pools, enabling more robust analyses across users but still offer limited sessions per user. While data recorded for the Cybathlon competitions \cite{jaeger2023cybathlon} addresses this limitation by including multiple sessions over extended periods \cite{perdikis2018cybathlon}, it focuses solely on a single patient, i.e., the pilot competing in the event. 
	The only other dataset comparable to the one used in this study is provided by \textit{Forenzo et al.} \cite{forenzo2024continuous}. However, it includes significantly fewer subjects and sessions than the Stieger2021 dataset. Moreover, since their study alternates between different decoders, they were not able to observe any user learning over time. While this potentially provides valuable insights into the impact of decoder stability on user learning, it raises questions about the dataset's suitability to investigate (mutual) learning dynamics.
	
	\textit{Stieger2021 dataset}: 
	The dataset was collected at a sampling rate of \SI{1000}{\hertz} using 64 EEG channels. Data collection spanned 7 to 11 sessions, with an 8-week gap between the first two sessions, followed by sessions recorded every 2-3 days. For our analysis, we selected 24 channels centered around the motor cortex, resampled the data to \SI{250}{\hertz}, and applied a band-pass filter between \SI{8}{\hertz} and \SI{30}{\hertz} to capture the $\mu$ and $\beta$ rhythm. 
	Each session consisted of 450 trials, equally divided among three paradigms: left/right movement, up/down movement, and combined 2D movement. 
	Participants in the \emph{left/right (LR)} movement paradigm imagined opening and closing their right (left) hand to move the cursor to the right (left). 
	In the \emph{up/down (UD)} movement paradigm, they imagined opening and closing both hands to move the cursor upward and voluntary rest to move it downward. 
	The combined 2D movement paradigm required participants to integrate both types of imagery for two-dimensional cursor control. 
	For this study, we focused exclusively on the binary paradigms (LR and UD), as the four-class task typically results in accuracies too low for practical control \cite{stieger2021continuous, stieger2021benefits}. 
	This results in 150 trials per session per paradigm.

	At the beginning of each trial, the target appeared on the screen for \SI{2}{\second}, indicating the desired direction of cursor movement. During the subsequent feedback phase, participants had up to \SI{6}{\second} to steer the cursor toward the target. 
	The trial ended earlier if any target, i.e., an edge of the screen, was reached. 
	The position of the cursor, serving as visual feedback, was determined using an autoregressive (AR) model of order 16, and was updated every \SI{40}{\milli\second}. 
	This AR model serves as the baseline for our investigations, representing a widely established online decoding method.
	
	\subsection{Model}
	We employ the BaseNet \cite{wimpff2024eeg} architecture, which can be considered a modern evolution of the shallow convolutional neural networks ShallowNet \cite{schirrmeister2017deep} and EEGNet \cite{lawhern2018eegnet}. To make the architecture suitable for online decoding, we use real-time adaptive pooling (RAP) \cite{wimpff2024tailoring}, which modifies the pooling layers to enable the decoding of sliding windows. We select sliding windows of length \SI{1}{\second} (as done in, e.g., \cite{forenzo2024continuous, wimpff2024tailoring, stieger2021benefits}) and an update frequency of \SI{25}{\hertz} to match the original online setting. This leads to the RAP parameters $k_1=s_1=5, k_2=50, s_2=2$. The complete source code is available on GitHub \footnote{https://github.com/martinwimpff/eeg-continual}.
	
	\subsection{Training strategies}
	\begin{figure}
		\includegraphics[width=\columnwidth]{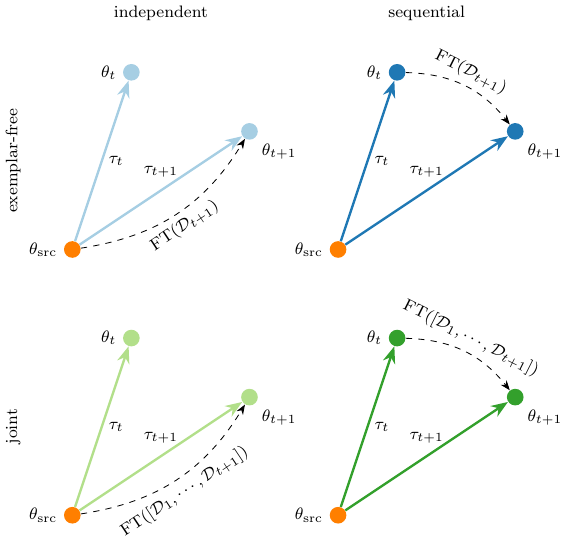}
		\caption{Fine-tuning process and task vectors across different settings in a two-dimensional weight space. Solid arrows represent task vectors, dashed lines illustrate the fine-tuning trajectory.}
		\label{fig:tvs}
	\end{figure}
	Each subject participates in up to 11 sessions, with each session containing $N_t = 150$ trials $X_i$ per paradigm paired with corresponding labels $y_i$. The dataset for a single session is represented as 
	$\mathcal{D}_{\text{session}}^{\text{subject}} = \{(X_i, y_i)\}_{i=1}^{N_t}$.
	During supervised pre-training, we employ a cross-subject leave-one-subject-out strategy to learn subject-invariant representations as done in \cite{ctrl2024generic}, providing an effective initialization for subsequent subject-specific fine-tuning. The source dataset is constructed by aggregating data from the first session of the remaining $N - 1 = 60$ subjects, denoted as
	$\mathcal{D}_{\text{source}}^{\text{i}} = \bigcup_{j \in \{1, \dots, N\} \setminus \{i\}} \mathcal{D}_1^j$. 
	This approach enables calibration-free online decoding for unseen subjects by leveraging data from the other participants. Moreover, it reflects a practical scenario in which only limited data from multiple subjects is available at the initial stage of source model training. Subsequent subject-specific applications can then be highly personalized.

	After pre-training, the model undergoes supervised fine-tuning using subject-specific data under a causal constraint, ensuring that only data recorded prior to the test session is used for fine-tuning.
	This fine-tuning step is essential as it enables the model to adapt to subject-specific patterns, which can vary widely across users. By refining the pre-trained, subject-invariant model with personalized data, the model becomes better suited to each user’s unique characteristics, resulting in improved decoding performance.
	
	We investigate two distinct data settings typically examined in continual learning: \emph{exemplar-free} and \emph{joint}. In the exemplar-free setting, only the data from the prior session is used for fine-tuning, whereas the joint setting incrementally incorporates data from all prior sessions.
	
	Additionally, we evaluate two fine-tuning strategies: \textit{independent} and \textit{sequential}. The independent strategy reinitializes fine-tuning from the pre-trained source model for each new session, while the sequential strategy builds upon the most recently fine-tuned model of the target subject. Combining both, there are four different settings in total: \textbf{\textcolor{examplefreeind} {exemplar-free independent}}, \textbf{\textcolor{examplefreeseq} {exemplar-free sequential}}, \textbf{\textcolor{joininde} {joint independent}} and \textbf{\textcolor{joinseq} {joint sequential}}, which are compared to the \textbf{\textcolor{baseline} {baseline}} from \cite{stieger2021continuous} and the non-adapted, subject-invariant \textbf{\textcolor{source} {source}} model.
	
	\subsection{Task vector notation}
	A widely used representation for describing the variations in fine-tuned models is the task vector notation \cite{ilharco2022editing}. For simplicity, we will omit the target subject index in the following explanation. Here, $\theta_{\text{src}}$ represents the model weights after pre-training on the source data, while $\theta_t$ denotes the weights following the $t$-th fine-tuning iteration. The task vector is defined as $\tau_t = \theta_t - \theta_{\text{src}}$, which specifies a direction within the weight space.
	The task vector notation of the four different settings together with the corresponding fine-tuning trajectory is visualized in Fig. \ref{fig:tvs}.

	Following the causal constraint of only using the previous sessions for fine-tuning, the fine-tuned weights $\theta_t$ are evaluated using the datset $\mathcal{D}_{t+1}$ from the session $t+1$. 
	
	\subsection{Test-time adaptation}
	As rebiasing the decoder between different domains \cite{wimpff2024tailoring, kumar2019towards} is a very important step in MI but recording additional calibration data for each new domain is costly \cite{rao2024once}, we perform online test-time adaptation (OTTA) \cite{li2023t, wimpff2024calibration}. Specifically, we use Euclidean alignment (EA) \cite{he2019transfer, junqueira2024systematic} in an online fashion \cite{wimpff2024tailoring} to mitigate the domain shift of the input data between sessions. 
	Additionally, we use online Adaptive batch norm (AdaBN) \cite{li2016revisiting, schneider2020improving} to account for the shifts in the batch normalization statistics in the intermediate layers of the model. 
	Both adaptation processes are carried out using only the current (sliding) window of the target session, making this a single-sample OTTA approach.
	The use of OTTA makes it possible to dispense domain-specific calibration data without losing performance. Another advantage is that, since as we only account for the overall distribution shift, the decision boundary does not change within one session, ensuring stability and thus facilitating user learning \cite{perdikis2018cybathlon}.
	
	\subsection{Metrics}
	
	To compare our different approaches, we use the trial-wise accuracy \cite{dreyer2023large} as our primary metric. This means that a trial is considered to be successful if more than \SI{50}{\percent} of all windows of that trial are correctly classified. For simplicity, we will refer to this as accuracy in the following. \newline
	Reported single values with standard deviations correspond to the mean and standard deviation of individual subject performances. This is achieved by first averaging sessions per subject, ensuring equal representation regardless of the number of sessions completed by each subject.
	
	To compare the similarity between task vectors, we employ the cosine distance $d(\tau_i,\tau_j) = 1 - \frac{\tau_i\cdot\tau_j}{||\tau_i||\cdot||\tau_j||}\in[0,1]$.
	
	\section{RESULTS \& DISCUSSION}
	
	\begin{figure*}[t]
		\centering
		\begin{subfigure}{0.49\textwidth}
			\centering
			\includegraphics[width=\linewidth]{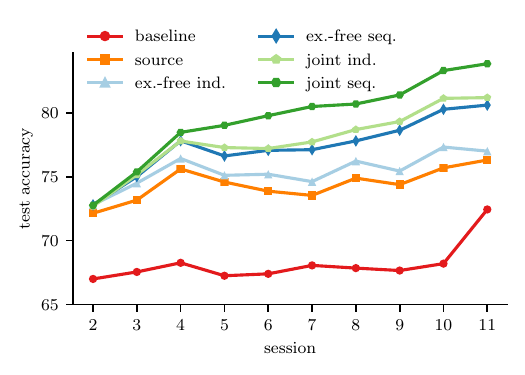}
			\caption{LR paradigm}
			\label{fig:session-LR}
		\end{subfigure}
		\hfill
		\begin{subfigure}{0.49\textwidth}
			\centering
			\includegraphics[width=\linewidth]{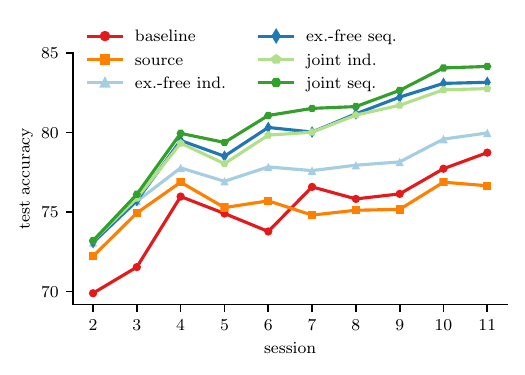}
			\caption{UD paradigm}
			\label{fig:session-UD}
		\end{subfigure}
		\caption{Accuracy across the sessions for each fine-tuning strategy compared to the source model and the baseline. Each dot represents the test accuracy of a single session, averaged across all subjects. The x-axis denotes session progression, while the y-axis represents the test accuracy (\%). 
		}
		\label{fig:session-comparison}
	\end{figure*}
	
	\begin{figure}
		\centering
		\includegraphics[width=\columnwidth]{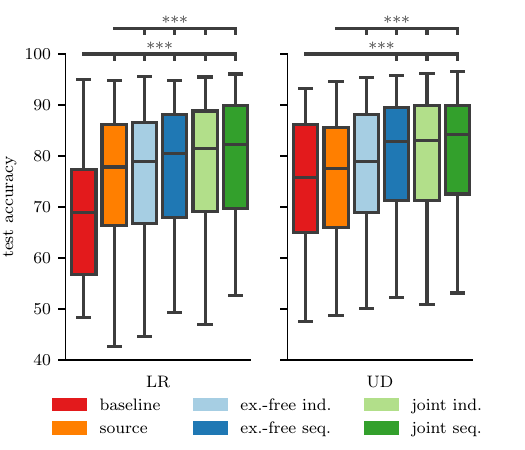}
		\caption{Average test accuracy over all sessions (2 - 11) per setting. Stars above the brackets indicate a significance level ($p<0.001$ (***)) when compared to the baseline and source model, respectively.}
		\label{fig:avg-perf}
	\end{figure}
	
	Figures \ref{fig:session-LR} and \ref{fig:session-UD} present the session-wise test accuracy for the different approaches corresponding to the LR and UD paradigms, respectively. 
	For the baseline and our source model, which remain constant (apart from rebiasing) across the sessions for each subject, the performance increase over time can only be attributed to the user exhibiting more discriminative patterns over time. It is, however, worth noting that as our experiments are carried out in a pseudo-online fashion \cite{carrara2024pseudo}, potential user adaptation to our decoders can not be explicitly examined. \newline
	Both figures clearly demonstrate that incorporating decoder adaptation enhances the average test accuracy and increases the extent of performance improvement over time. This trend is indicative of successful decoder adaptation. 
	We speculate that in an online experiment, the improvement over time could be even greater, as users would have the opportunity to adjust to the adapting decoder. 
	
	The average performance, together with the pairwise p-values (paired two-sided t-test against baseline and source model), are displayed in Fig. \ref{fig:avg-perf}. 
	For the LR paradigm, all our models outperform the baseline, and all fine-tuning approaches outperform the source model with $p<0.001$. For the UD paradigm, the source and baseline are not statistically different ($p=0.748$), and the difference between the baseline and the exemplar-free independent setting is smaller ($p=0.0124$).
	
	These findings confirm the decoder adaptation's effectiveness while highlighting key differences between the approaches. Notably, leveraging previously acquired subject-specific knowledge, whether explicitly through joint fine-tuning or implicitly by using the previously fine-tuned decoder, leads to improvements in overall performance. 
	
	The joint sequential setting achieves the highest accuracy, as it benefits from cumulative progress by reusing the previously fine-tuned model, in contrast to the independent approach. Furthermore, this setting ensures stable fine-tuning by incorporating all previously recorded subject-specific data, unlike the exemplar-free approach.
	
	\begin{figure}
		\centering
		\includegraphics[width=\columnwidth]{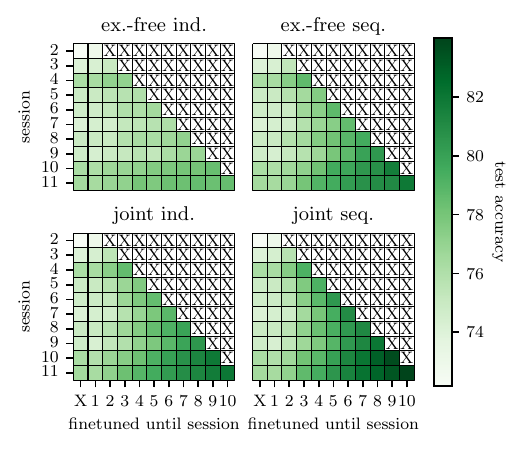}
		\caption{Average test accuracy (over all subjects and both paradigms) for each fine-tuning setting. The X on the x-axis refers to no fine-tuning, i.e., using the source model.}
		\label{fig:eval-ft}
	\end{figure}
	
	To further understand the relationship between the fine-tuning approaches, we visualize the session-wise test accuracy for each fine-tuned model in Fig. \ref{fig:eval-ft}.  In each matrix, the first column represents the source performance, while the diagonal entries on the right correspond to the fine-tuning results reported in Fig. \ref{fig:session-comparison}. Since we adhere to a strictly causal data setting, sessions are not evaluated using models fine-tuned with data recorded in later sessions. Thus, these entries are marked with an X.
	
	For three out of four settings, we observe a clear trend: performance improves with both the progression of sessions (user learning) and increased fine-tuning steps (decoder adaptation). When comparing rows, selecting the most recently fine-tuned model (i.e., the field on the right) exhibits the highest test accuracy. 
	
	In the exemplar-free independent scenario, these trends are still apparent but show a diminished degree of improvement across sessions and fine-tuning steps. We attribute this to reduced stability, as fine-tuning relies only on data from the most recent session, which can vary significantly due to factors such as user concentration, motivation, or environmental influences. Consequently, a single session may not adequately represent the current state of the user.
	Moreover, in this scenario, the decoder cannot leverage previously acquired subject-specific knowledge and must begin adapting to the user from scratch for each new session, limiting its ability to achieve cumulative progress. 
	
	The displayed matrices represent averages across subjects and paradigms, and subject-specific matrices might deviate from the overall trend. Thus, to further assess the validity of selecting the most recently fine-tuned decoder, we conducted a theoretical experiment. For each subject, we identified the best-performing decoder per session to establish a theoretical upper bound. 
	For Fig. \ref{fig:eval-ft}, this corresponds to picking the highest value per row in each subject-specific matrix.

	\begin{table}[!htbp]
		\caption{Theoretical upper bounds for each setting and paradigm, together with the performance of the most recently fine-tuned decoder.}
		\begin{center}
			\resizebox{\linewidth}{!}{
				\begin{tabular}{c|cc|cc}
					\toprule
					\multirow{2}{*}{\bf setting} & \multicolumn{2}{c|}{\bf LR} & \multicolumn{2}{c}{\bf UD} \\
					\cmidrule{2-5}
					& \textbf{most recent} & \textbf{upper bound} & \textbf{most recent} & \textbf{upper bound} \\
					\midrule
					ex.-free ind. & $75.1\pm13.6$ & $76.7\pm13.3$ & $77.1\pm12.6$ & $78.8\pm12.4$ \\
					ex.-free seq. & $76.9\pm13.2$ & $78.0\pm12.9$ & $79.1\pm12.3$ & $80.2\pm12.1$ \\
					joint ind. & $77.2\pm13.4$ & $77.9\pm13.2$ & $78.8\pm12.8$ & $79.9\pm12.4$ \\
					joint seq. & $78.8\pm12.6$ & $79.5\pm12.4$ & $79.8\pm12.3$ & $80.9\pm12.0$ \\
					\bottomrule
			\end{tabular}}
			\label{tab:upper}
		\end{center}
	\end{table}

	These upper bounds are presented in Table \ref{tab:upper} for each setting and paradigm, together with the value for picking the most recently fine-tuned decoder. P-values are omitted, as all comparisons are statistically significant with $p<0.001$.
	As expected, our strategy performs worse than the theoretical upper bound. However, the overall performance difference is surprisingly small, except for the exemplar-free independent setting. 
	Encouragingly, this suggests that the strategy of selecting the most recently fine-tuned model closely approximates the theoretical optimal decoder selection.
	In the exemplar-free setting, which still has the lowest upper bound among the settings, adaptation is less stable, making it beneficial to switch between different stages of fine-tuning.

	\begin{figure}
		\includegraphics[width=\columnwidth]{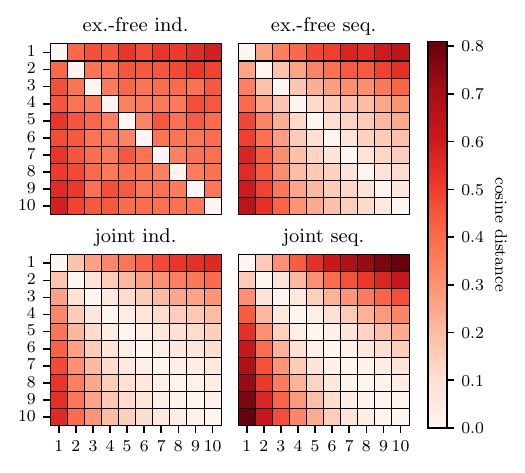}
		\caption{Cosine distance between task vectors for all strategies.}
		\label{fig:tv-dist}
	\end{figure}
	To examine the stability between fine-tuned models more closely, we calculated the cosine distance between task vectors, as shown in Fig. \ref{fig:tv-dist}. 
	The distance matrix for the exemplar-free independent scenario supports our earlier conclusions. Since fine-tuning restarts from scratch for each session, the process is less stable, resulting in larger distances between consecutive task vectors compared to the other three settings. Nevertheless, the cosine distance remains well below 1, indicating a degree of relationship and some level of stability between consecutive task vectors. In contrast, for task- or class-incremental fine-tuning, the cosine distance approaches one \cite{ilharco2022editing}, signifying nearly orthogonal task vectors.

	For the other three settings, especially the later task vectors exhibit greater similarity to one another, suggesting increased stability between fine-tuning steps, which tends to benefit user learning \cite{forenzo2024continuous, perdikis2018cybathlon}. Additionally, across all four settings, the distance to the first task vector is noticeably larger. This may be attributed to greater data discrepancies, given the 8-week gap between the first two sessions compared to only a few days between each of the subsequent ones \cite{stieger2021continuous}.
	
	\begin{table}[htbp]
		\caption{Ablation study for the number of previous sessions used during sequential fine-tuning.}
		\begin{center}
			\resizebox{\linewidth}{!}{
				\begin{tabular}{c|cc|cc}
					\toprule
					\multirow{2}{*}{\bf \# previous sessions} & \multicolumn{2}{c|}{\bf LR} & \multicolumn{2}{c}{\bf UD} \\
					\cmidrule{2-5}
					& \textbf{accuracy} & \textbf{p-value} & \textbf{accuracy} & \textbf{p-value} \\
					\midrule
					joint & $78.8\pm12.6$ & $X$ & $79.8\pm12.3$ & $X$ \\ \midrule
					4 & $78.7\pm12.5$ & $0.038$ & $80.0\pm12.1$ & $0.036$ \\
					3 & $78.3\pm12.7$ & $<0.001$ & $80.0\pm12.1$ & $0.231$ \\
					2 & $77.7\pm12.8$ & $<0.001$ & $79.8\pm12.1$ & $0.974$ \\ 
					exemplar-free & $76.9\pm13.2$ & $<0.001$ & $79.1\pm12.3$ & $0.013$ \\
					\bottomrule
			\end{tabular}}
			\label{tab:ablation-buffer}
		\end{center}
	\end{table}
	
	\begin{table}[htbp]
		\caption{OTTA ablation study for the source model. P-values (paired two-sided t-test) calculated against the first row.}
		\begin{center}
			\resizebox{\linewidth}{!}{
				\begin{tabular}{cc|cc|cc}
					\toprule
					\multirow{2}{*}{\bf EA} & \multirow{2}{*}{\bf AdaBN} & \multicolumn{2}{c|}{\bf LR} & \multicolumn{2}{c}{\bf UD} \\
					\cmidrule{3-6}
					& & \textbf{accuracy} & \textbf{p-value} & \textbf{accuracy} & \textbf{p-value} \\
					\midrule
					\cmark & \cmark & $74.0\pm 13.8$ & $X$ & $75.2\pm13.0$ & $X$ \\  
					\xmark & \cmark & $64.3\pm13.5$ & $<0.001$ & $64.9\pm10.5$ & $<0.001$ \\  
					\cmark & \xmark & $71.0\pm13.1$ & $0.061$ & $69.3\pm15.0$ & $<0.001$ \\  
					\xmark & \xmark & $57.9\pm11.2$ & $<0.001$ & $59.9\pm9.90$ & $<0.001$ \\  
					\bottomrule
			\end{tabular}}
			\label{tab:ablation}
		\end{center}
	\end{table}
	
	The previously presented results demonstrate the advantages of training models in a joint sequential manner. However, since the joint data setting increases the memory and time requirements for fine-tuning, we examined how the number of previous sessions used during sequential fine-tuning influences performance. The results are shown in Table \ref{tab:ablation-buffer}, with p-values calculated against the joint setting (first row). 
	For the LR paradigm, performance differences are significant across all data settings, though a trend emerges where accuracy approaches the joint setting when incorporating four previous sessions. In contrast, for the UD paradigm, the differences are generally smaller. Interestingly, fine-tuning with only three or four previous sessions slightly outperforms the joint setting. We speculate that this may be due to greater user learning over time (see Figure \ref{fig:session-UD} baseline), where older sessions could hinder rapid adaptation to new data.
	Nonetheless, the differences are still small, and this hypothesis is speculative as we can not examine the user's behavior in the different settings. Therefore, we still recommend using the joint setting for its stability. However, if memory constraints or a large number of sessions become a concern, this setting could likely be relaxed without a (large) loss in performance.
	
	To eliminate the need for session-specific calibration data while maintaining performance, we utilize OTTA throughout this study, which integrates online EA and online AdaBN. To evaluate the individual contributions of these components, we conducted an ablation study using the source model. The results are presented in Table \ref{tab:ablation}, where p-values are calculated against the fully enabled configuration (first row).
	
	Notably, disabling EA alone contributes to a $\sim10\%$ loss in performance compared to the source model with both components active. 
	Similarly, disabling AdaBN also decreases performance, though its statistical significance ($p<0.05$) is only observed for the UD paradigm. 
	This may be due to the generally higher level of user learning observed in the UD paradigm, meaning the impact of OTTA is more pronounced when the distribution shift is greater.
	Ultimately, while EA has a more substantial effect, both components play a crucial role in ensuring reliable and robust decoding performance.
	
	\section{CONCLUSION}
	This study explored various fine-tuning strategies for pseudo-online longitudinal EEG MI decoding in a causal setting for a large user group. We investigated the impact of different training strategies to incorporate previously acquired subject-specific knowledge implicitly or explicitly during fine-tuning with various training designs. 
	The results demonstrate that leveraging previously acquired subject-specific information enhances performance and improves stability. Overall, the most effective strategy, \emph{joint sequential} fine-tuning, involved incorporating all previously recorded subject-specific data while continuously building on the last subject-specific fine-tuned model.\newline
	Furthermore, the use of OTTA enables calibration-free operation for new sessions or subjects, making our approach well-suited for real-world applications and directly integrable into both applicative and experimental contexts.\newline
	Although these results are obtained through offline analysis, we make every effort to replicate a pseudo-online setting that closely approximates real conditions. While our experimental analysis only accounts for one half of the \emph{two-learner} system, we are confident that these results are robust and will hold in online experiments as this study is conducted on a large user group and across several sessions.

	\addtolength{\textheight}{-1.4cm}   
	


	
	
	\newpage
	\bibliographystyle{ieeeconf}  
	\bibliography{bib}

\end{document}